\documentclass[preprint,12pt,authoryear]{elsarticle}

\usepackage{amssymb}
\usepackage{amsmath}

\usepackage{lineno}

\usepackage{xcolor}
\definecolor{newcolor}{rgb}{.8,.349,.1}
\usepackage[pagebackref=False,breaklinks=true,colorlinks,bookmarks=false]{hyperref}
\usepackage{soul}
\usepackage[utf8]{inputenc}
\usepackage[small]{caption}
\usepackage{multirow}
\usepackage{comment}
\usepackage{booktabs}
\usepackage{graphicx}
\usepackage{graphics}
\usepackage[export]{adjustbox}
\usepackage{bm}
\usepackage{amsmath,amsfonts}
\usepackage{subcaption }

\def\eg{\emph{e.g.}} 
\def\ie{\emph{i.e.}}

\journal{Knowledge-Based Systems}

\begin{document}

\begin{frontmatter}



\title{Collaborative Group: Composed Image Retrieval \\ via Consensus Learning from Noisy Annotations} 


\author[1]{Xu Zhang} 
\author[2]{Zhedong Zheng \corref{cor1}}
\ead{zhedongzheng@um.edu.mo}
\author[1]{Linchao Zhu}
\author[1]{Yi Yang}
\cortext[cor1]{Corresponding author.}


\affiliation[1]{organization={College of Computer Science and Technology, Zhejiang University},
            city={Hangzhou},
            postcode={310058}, 
            country={China}}

\affiliation[2]{organization={Faculty of Science and Technology, and Institute of Collaborative Innovation, University of Macau},
            city={Macau},
            postcode={999078}, 
            country={China}}

\begin{abstract}
Composed image retrieval extends content-based image retrieval systems by enabling users to search using reference images and captions that describe their intention. 
Despite great progress in developing image-text compositors to extract discriminative visual-linguistic features, we identify a hitherto overlooked issue, triplet ambiguity, which impedes robust feature extraction. \textcolor{black}{Triplet ambiguity refers to a type of semantic ambiguity that arises between the reference image, the relative caption, and the target image. It is mainly due to the limited representation of the annotated text, resulting in many noisy triplets where multiple visually dissimilar candidate images can be matched to an identical reference pair (i.e., a reference image + a relative caption).} 
To address this challenge, we propose the Consensus Network (Css-Net), inspired by the psychological concept that groups outperform individuals. Css-Net comprises two core components: (1) a consensus module with four diverse compositors, each generating distinct image-text embeddings, fostering complementary feature extraction and mitigating dependence on any single, potentially biased compositor; (2) a Kullback-Leibler divergence loss  that encourages learning of inter-compositor interactions to promote consensual outputs.
During evaluation, the decisions of the four compositors are combined through a weighting scheme, enhancing overall agreement. On benchmark datasets, particularly FashionIQ, Css-Net demonstrates marked improvements. Notably, it achieves significant recall gains, with a 2.77\% increase in R@10 and 6.67\% boost in R@50, underscoring its competitiveness in addressing the fundamental limitations of existing methods.

\end{abstract}


\end{frontmatter}


\section{Introduction}
\label{sec:intro}
\begin{figure}[ht!]
    \centering
    \includegraphics[width=0.85\linewidth]{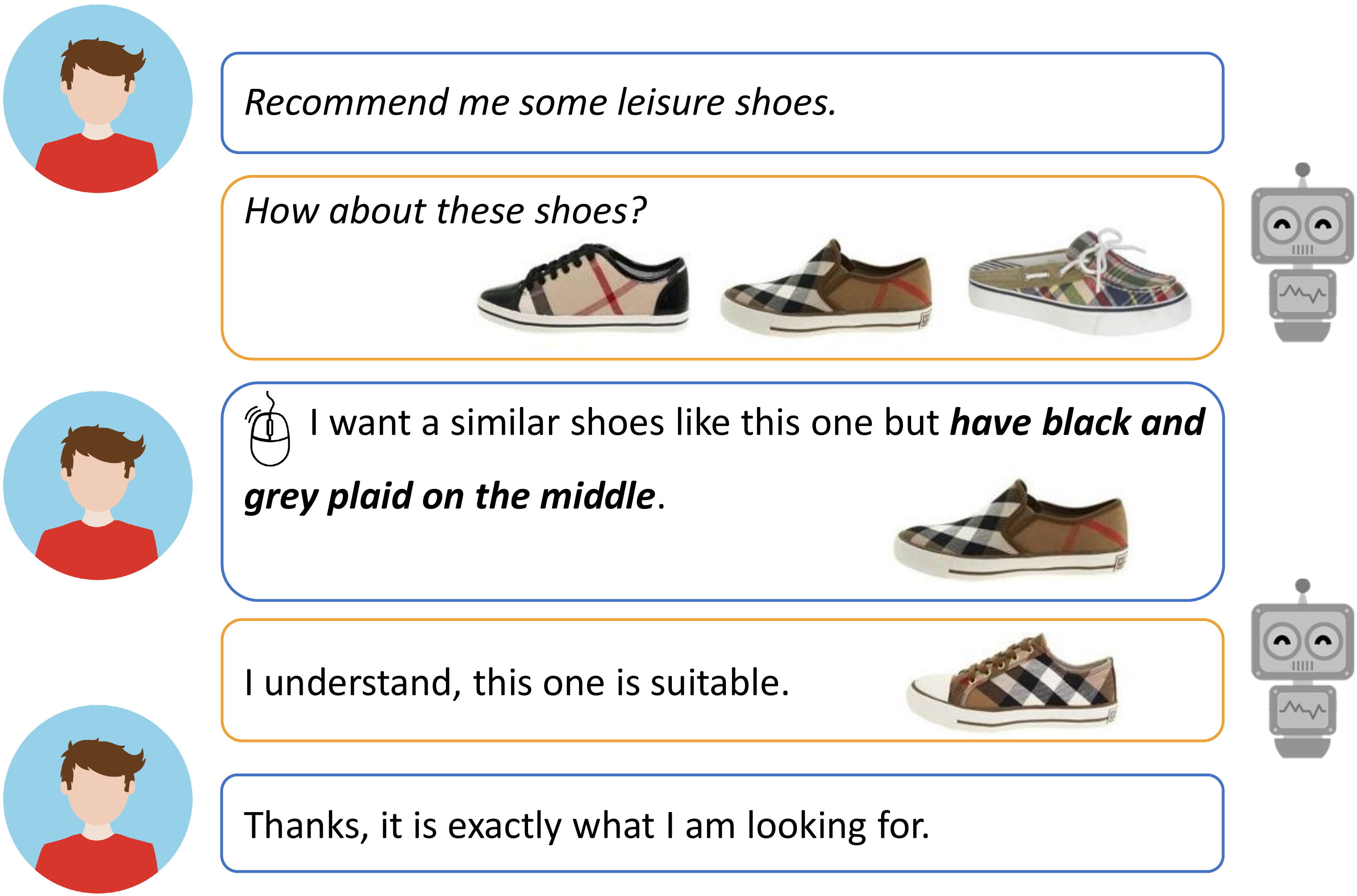}
    \caption{Schematic illustration of the composed image retrieval system. Through using a reference image and a relative caption, the system endeavors to precisely retrieve the intended target image from all candidate images.}
    \label{fig:task}
\end{figure}

Image retrieval plays a pivotal role in computer vision and proves to be valuable in many applications, such as product search\!~\citep{guo2019attentive,sharma2019retrieving,guo2018multi}, internet search\!~\citep{noh2017large} and fashion retrieval\!~\citep{liu2016deepfashion,liao2018interpretable}. Prevalent image retrieval approaches include image-to-image retrieval\!~\citep{deng2019arcface,fan2019spherereid,sheng2020mining,hafner2022cross} and text-to-image retrieval\!~\citep{zhen2019deep,zheng2020dual,guerrero2021cross,wang2022point}, which endeavor to locate the image of interest using a single image or descriptive texts as a query. 
Despite significant progress, users often lack a precise search target in advance but instead seek categories, such as shoes or clothing. Therefore, an interactive system is highly desirable to assist users to reconsider their intentions, as depicted in Fig.\!~\ref{fig:task}. Hence, Composed image retrieval, which aims to search the image of interest given the composed query consisting of a reference image and a relative caption describing the modification, has attracted great attention\!~\citep{vo2019composing,chen2020image,lee2021cosmo,kim2021dual,wen2021comprehensive}.

Recent studies addressing the task of composed image retrieval primarily concentrate on extracting discriminative representations from image-text-image triplets. For example, TIRG\!~\citep{vo2019composing}, VAL\!~\citep{chen2020image}, and CoSMo\!~\citep{lee2021cosmo} propose different ways to modify the visual features of the reference image conditioned on the relative caption. TIRG uses a simple gating and residual module, VAL devises a visual-linguistic attention learning framework, and CoSMo introduces the content and style modulators. Additionally, CLVC-Net\!~\citep{wen2021comprehensive} and CLIP4cir\!~\citep{baldrati2022conditioned} devise more intricate multi-modal fusion modules to accentuate the modifications of the reference image. CLVC-Net uses local-wise and global-wise compositors, while CLIP4cir finetunes the CLIP\!~\citep{radford2021learning} text encoder and trains a combiner network to fuse features.

Despite the significant success, these works fail to address an inherent problem of the composed image retrieval task: the ambiguity of the training data triplets, \ie, \textbf{triplet ambiguity}. Triplet ambiguity originates from the annotation process where annotators focus on single data triplet, and frequently describe simple properties such as color and size, while neglecting more fine-grained details, such as location and style. Consequently, many noisy triplets exist where candidate images meet the requirement of the composed query but are not annotated as the desired ground-truth target image, especially when the relative caption is brief. Similar annotation ambiguity is also observed in pair-wise data\!~\citep{wray2021semantic,falcon2022feature} and remains challenging. \textcolor{black}{As shown in Fig.\!~2, existing methods treat composed image retrieval as an instance-level retrieval, that is, given a reference pair (comprising a reference image and a relative caption), only the annotated target image is considered as the correct image to retrieve. In fact, due to the limitation of the text description, many candidate images within the dataset are semantically similar to the point of being identical, but are treated as the negative counterparts, thus producing many noisy triplets.} These noisy triplets  compromises the representation learning of the single compositor, since the metric learning objective in this task aims to push away these false-negative samples from the composed query. Empirically, we verify the existence of triplet ambiguity in Sec.~\!\ref{subsec:ambiguity}.

\begin{figure}[t]
    \centering
    \includegraphics[width=.9\linewidth]{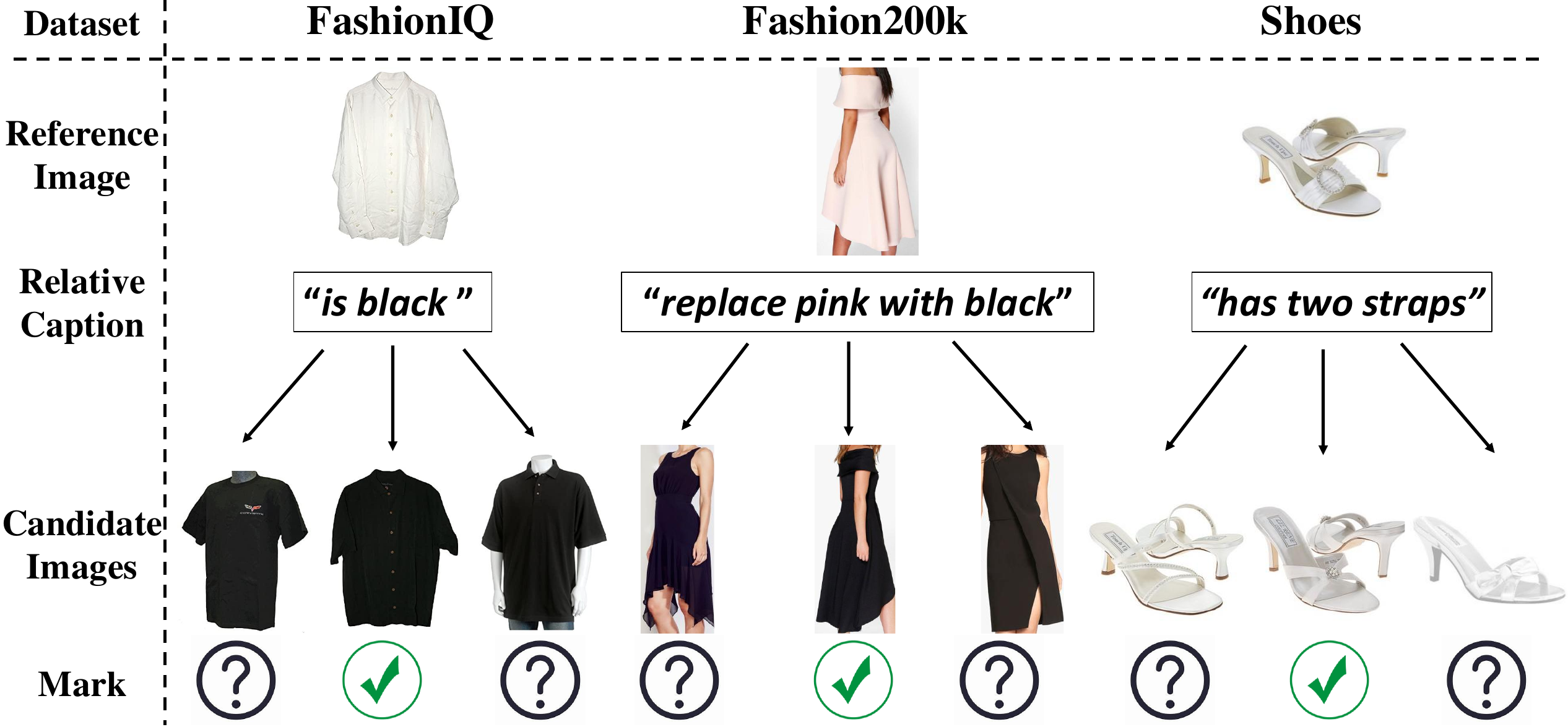}
    \caption{Illustration of the triplet ambiguity problem. Triplet ambiguity denotes multiple false-negative samples in the dataset as the annotator usually see one triplet with true match (\protect\includegraphics[scale=0.15,valign=c]{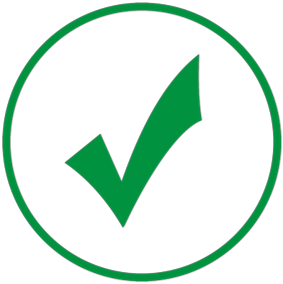}) at a time, while neglecting other candidates (\protect\includegraphics[scale=0.15,valign=c]{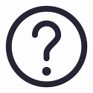}). 
    }
    \label{fig:intro}
\end{figure}

To relieve the triplet ambiguity problem, we propose a straightforward and effective Consensus Network (Css-Net) for composed image retrieval, as illustrated in Fig.~\!\ref{fig:overview}(a). 
The key idea underpinning our method to alleviate the triplet ambiguity is ``two heads are better than one'' in short. To be more specific, an individual often errs due to the biases caused by noisy triplets, but groups are less susceptible to making similar mistakes, thereby circumventing sub-optimal solutions. This is known as the psychological finding\!~\citep{hinsz1990cognitive} that groups perform better than individuals on the memory task. Consequently, we aim to (1)~\!develop a consensus module (group) composed of compositors possessing diverse knowledge to jointly make decisions during evaluation and (2)~\!encourage learning among different compositors to minimize their biases learned on noisy triplets by employing an additional Kullback Leibler divergence loss (KL loss)\!~\citep{kullback1951information}. 

Css-Net ensures that the compositors possess distinct knowledge in two ways: $\bullet$~Motivated by the finding\!~\citep{lin2017feature,miech2021thinking} that the image features of high-resolution are semantically weak, while the image features of low-resolution are semantically strong, we employ two image-text compositors at different depths of the same image encoder, (\ie, block3 and block4 of the ResNet\!~\citep{he2016deep}). The former focuses more on detailed change like ``has a purple star pattern'', while the latter emphasizes more overall change such as ``is modern and fashional''. $\bullet$~Unlike the image-text compositor that uses relative caption to describe \textbf{what to change} on the reference image, we devise the text-image compositor to capture the textual cues based on text-to-image retrieval, where the reference image implies \textbf{what to preserve} for the referenece image. See details in Sec.\!~\ref{subsec:overview}. 
To minimize the negative impact of triplet ambiguity during training, we impose a KL loss between two image-text compositors. The KL loss promotes two compositors to learn from each other and reach a consensus, which is similar to supervision from peers in a group, as it helps each compositor to reduce its own bias and thus avoids overfitting to the annotated target image. 

In summary, our contributions are as follows:

    $\bullet$ 
    We have identified an inherent issue within the context of composed image retrieval, namely triplet ambiguity, which we subsequently confirm through initial experimental investigations (\textit{see Fig.\!~\ref{fig:intro}~and\!~\ref{fig:twolosses}}).  This problem, stemming from the inherent noisiness of the annotation process, results in suboptimal model learning, as it compromises the extraction of discriminative features that integrate visual and linguistic information.
    
    $\bullet$ To relieve triplet ambiguity, we introduce the Consensus Network (Css-Net) featuring a consensus module with four distinct compositors for collaborative training (\textit{see Table\!~\ref{tab:diagnostic}}) and joint inference (\textit{see Table\!~\ref{tab:joint_infer}}).
    
    $\bullet$ Extensive experiments show that the proposed method minimizes the negative impacts of noisy triplets. On three prevalent public benchmarks, we observe that Css-Net significantly surpasses the current state-of-the-art competitive methods, \eg, with $+2.77\%$ Recall@10 on Shoes, and $+6.67\%$ Recall@50 on FashionIQ (\textit{see Table\!~\ref{tab:fashioniq},\!~\ref{tab:shoes},~and\!~\ref{tab:fashion200k}}).

\section{Related Work}

\noindent\textbf{Cross-modal Image Retrieval.}
Cross-modal image retrieval has attracted wide attention from researchers. The most popular patterns of image retrieval are image-to-image matching\!~\citep{zheng2017discriminatively,deng2019arcface,sun2020circle,wu2017RGB,dai2018cross,liu2022learning,qu2024source} and text-to-image matching\!~\citep{liu2019modality,zhang2020context,liu2022image,zhang2022vldeformer,li2024integrating}. Although these paradigms have made great progress, they do not provide enough convenience for users to express their search intention. Therefore, more forms of image retrieval with flexible queries such as sketch-based image retrieval\!~\citep{deng2020progressive,wang2021tcn,li2022zero,liang2024sketch} have emerged. In this work, the composed image retrieval task involves a composed query of a reference image and a relative caption. To tackle this task, recent works\!~\citep{vo2019composing,chen2020image,yang2021cross,zhang2021heterogeneous,lee2021cosmo,wen2021comprehensive,gu2021image,zhao2022progressive,han2023fame} devise diverse composition architectures to capture the visual-linguistic relation. Unlike the methods described above, our Css-Net does not propose complicated compositors. Instead, our work mainly focuses on reducing single compositor biases to alleviate the identified triplet ambiguity problem.

\noindent\textbf{Attention Mechanism.}
The attention mechanism is widely used in language and vision tasks in machine learning to capture the relations between features. This mechanism is also inspired by a psychological finding\!~\citep{corbetta2002control} that humans observe and pay attention to specific parts as needed. In the composed image retrieval task, many works use the attention mechanism to design the image-text compositor. For example, VAL\!~\citep{chen2020image} employs self-attention to capture the image-text relations by concatenating the text feature to the image feature. CoSMo\!~\citep{lee2021cosmo} adopts the disentangled multi-modal non-local block to stabilize the training procedure for learning better representations. Besides, CLVC-Net\!~\citep{wen2021comprehensive} proposes a cross attention between each word in the sentence and each spatial location of the image feature to recognize details. 
In our work, the main idea is not to design a new attention-based compositor but to utilize several compositors to form as a consensus module. \textcolor{black}{Without loss of generality, we deploy the widely-used CoSMo as the image-text compositor. Moreover, we propose specific text-image compositors based on cross attention to better capture the relation between the reference image feature and the word-level text feature, which is orthogonal with existing attention-based models and could further improve the retrieval performance.}

\noindent\textbf{Co-training.}
Co-training is a semi-supervised learning technique that exploits two components to acquire complementary information on two views of the data\!~\citep{blum1998combining}. It has been extensively utilized in various research fields such as image recognition\!~\citep{qiao2018deep}, segmentation\!~\citep{peng2020deep,hui2023language} and domain adaptation\!~\citep{saito2018maximum,zheng2019unsupervised,luo2019taking}. Our work adopts a co-training paradigm that leverages four compositors with different knowledge to jointly make decisions for the composed image retrieval task.
The two image-text compositors focus on the detailed and overall changes to the reference images based on the perspective of finding ``what to change'' in the reference image, and the two text-image compositors are in view of the text-to-image retrieval with the reference image implying ``what to preserve'' for the relative caption. The compositors hold diverse knowledge from different views of the data. Thus, we explicitly encourage the consensus between compositors and leverage the consensus to rectify the single prediction.

\section{Method} \label{method}


\begin{figure*}[ht]
    \centering
    \includegraphics[width=\linewidth]{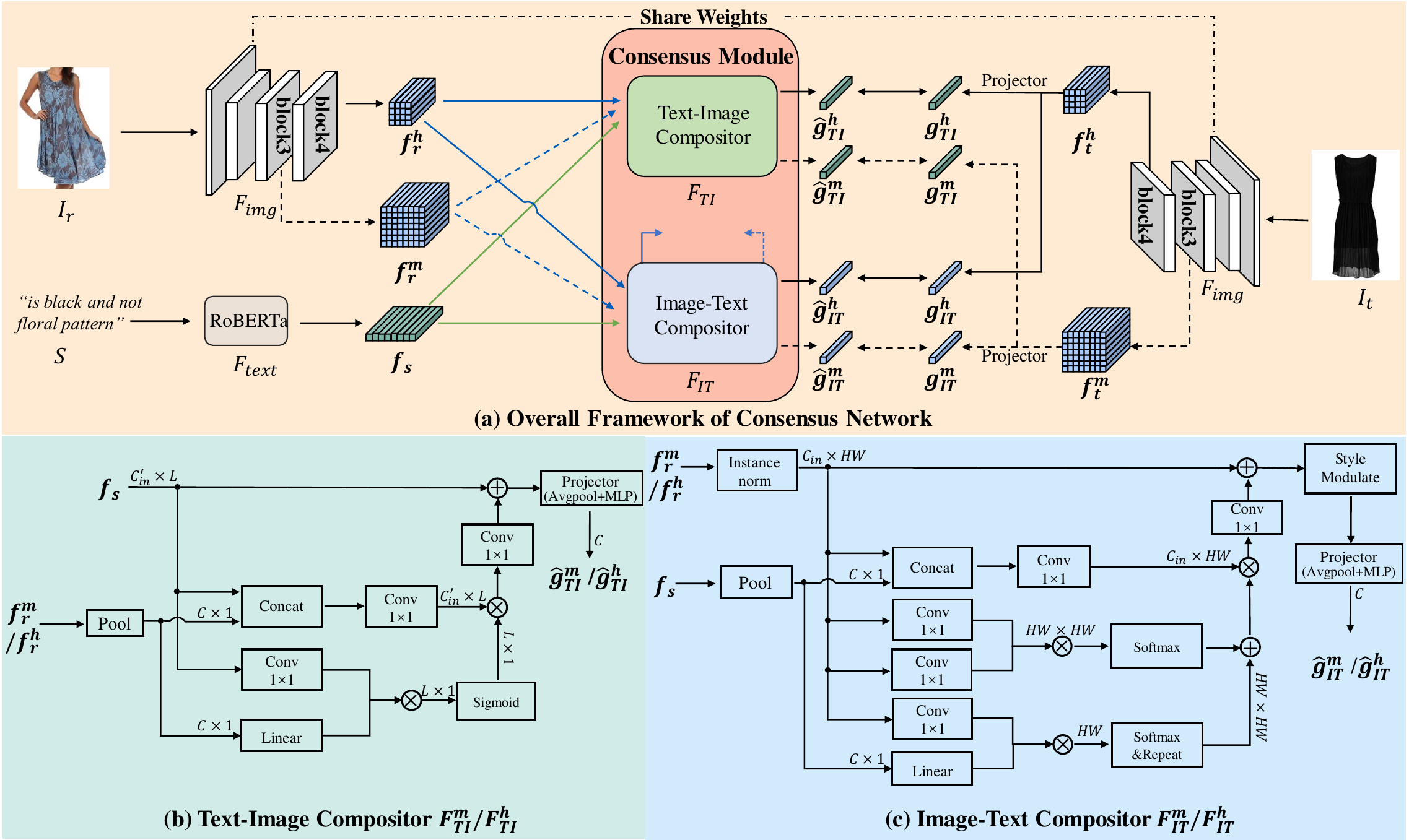}
    \put(-203, 166.5){\fontsize{7pt}{1em}\selectfont $\mathcal{L}_{\scriptstyle KL}$}
    \put(-150.5, 140.5){\fontsize{7}{14}\selectfont $\mathcal{L}^{\scriptstyle m}_{ IT}$}
    \put(-150.5, 158.5){\fontsize{7}{14}\selectfont $\mathcal{L}^{\scriptstyle h}_{IT}$}
    \put(-150.5, 191.5){\fontsize{7}{14}\selectfont $\mathcal{L}^{\scriptstyle m}_{TI}$}
    \put(-150.5, 210.5){\fontsize{7}{14}\selectfont $\mathcal{L}^{\scriptstyle h}_{TI}$}
    \caption{Schematic illustration of the Consensus Network. Given a reference image and a relative caption, the image encoder $F_{img}$ extracts the mid-level image feature $\bm{f_r^m}$ and high-level image feature $\bm{f_r^h}$, and the text encoder $F_{text}$ extracts the text feature $\bm{f_s}$. Then, compositors fuse the text feature with either the mid-level or high-level image feature. 
    Each compositor generates distinct composed feature. Finally, we match the composed features with the corresponding target features and impose a KL loss between image-text compositors for training.}
    \label{fig:overview}
\end{figure*}

\subsection{Overview of Consensus Network}
\label{subsec:overview}
  
As illustrated in Fig.\!~\ref{fig:overview} (a), the Consensus Network consists of three components: a image encoder, a text encoder, and a consensus module. The image encoder, $F_{img}$, extracts mid-level and high-level reference image features as $\bm{f^m_r}, \bm{f^h_r} = F_{img}(I_r),$ where $I_r$ is the reference image, and $\bm{f^m_r}, \bm{f^h_r} \in \mathbb{R}^{C_{in} \times (H \times W)}$ are mid-level and high-level image features, respectively (\ie, output from block3 and block4 of the ResNet\!~\citep{he2016deep}). 
$C_{in} \times (H \times W)$ represents the shape of the feature maps. For brevity, we do not distinguish between different shapes of image features. 
The text encoder, denoted as $F_{text}$, extracts features of the relative caption as $ \bm{f_s} = F_{text}(S),$ where $S$ denotes the relative caption, $\bm{f_s} \in \mathbb{R}^{C_{in}' \times L}$ refers to the word-level representation, and $L$ is the number of words of the relative caption.

After extracting the image and text features, the consensus module transforms the reference image features with the corresponding text features into the composed features. It consists of four distinct compositors possessing different knowledge. These compositors at different depths of the image encoder can be grouped into two types. Specifically, given the reference image feature $\bm{f_r}$ and the text feature $\bm{f_s}$, the composed query $\bm{\hat{g}}$ can be obtained by either an image-text compositor or a text-image compositor. 
The image-text compositor has the residual form of $\bm{\hat{g}_{IT}} = \bm{f_r} + comp(\bm{f_r},\bm{f_s})$ and mainly focuses on ``what to change'' for $\bm{f_r}$ conditioned on the relative caption, while the text-image compositor has the residual form of $\bm{\hat{g}_{TI}} = \bm{f_s} + comp(\bm{f_s},\bm{f_r})$ and mainly emphasizes on ``what to preserve'' for $\bm{f_s}$ conditioned on the reference image.
Here, $comp$ represents a trained component to fuse $\bm{f_r}$ and $\bm{f_s}$ as the condition. Considering both the performance and computational efficiency, the text-image compositors $F_{TI}^{m}$ and $F_{TI}^{h}$, shown in Fig.\!~\ref{fig:overview} (b), take the word-level representation $\bm{f_s}$ along with the average pooled reference image features $pool(\bm{f^m_r}), pool(\bm{f^h_r})$ as input, respectively:
\begin{equation}
\label{eq:text_comp}
    \begin{cases}
    &\bm{\hat{g}^m_{TI}} = F_{TI}^{m}(\bm{f_s}, pool(\bm{f^m_r})) \\
    &\bm{\hat{g}^h_{TI}} = F_{TI}^{h}(\bm{f_s}, pool(\bm{f^h_r})),
    \end{cases}
\end{equation}
where $\bm{\hat{g}^m_{TI}},~\bm{\hat{g}^h_{TI}}$ are the composed features from text-image compositors. Similarly, the image-text compositors $F_{IT}^m, F_{IT}^h$, shown in Fig.\!~\ref{fig:overview} (c) take the intermediate image feature maps, $\bm{f_r^m}, \bm{f_r^h}$ along with the pooled sentence-level text representation $pool(\bm{f_s})$ as input, which are given by:
\begin{equation}
\label{eq:img_comp}
    \begin{cases}
    &\bm{\hat{g}_{IT}^m} = F_{IT}^m(\bm{f^m_r}, pool(\bm{f_s})) \\
    &\bm{\hat{g}_{IT}^h} = F_{IT}^h(\bm{f^h_r}, pool(\bm{f_s})),
    \end{cases}
\end{equation}
where $\bm{\hat{g}_{IT}^m},~\bm{\hat{g}_{IT}^h}$ are the composed features from image-text compositors.

The target image features $\bm{f^m_t}, \bm{f^h_t}$ are obtained from the same image encoder $F_{img}$ as the reference image features $\bm{f^m_r}, \bm{f^h_r}$. Then four independent projector blocks (composed of an average pooling layer and a MLP) are employed to acquire target features: $\bm{g_{TI}^m}$, $\bm{g_{TI}^h}$, $\bm{g_{IT}^m}$, and $\bm{g_{IT}^h}$. Finally, the four compositors are trained by pulling close the corresponding target while pushing away other negatives within the embedding space.

\subsection{Consensus Module}
\label{subsec:consensus}

To relieve the triplet ambiguity, we introduce the consensus module, which consists of four distinct compositors with different knowledge. These compositors have individual biases learned on noisy triplets, which are minimized at two stages. At the training stage, each compositor acquires information from different views of the data, and the KL loss enables them to learn from each other to minimize biases. At the evaluation stage, each compositor independently provides decisions and collaborates to rank the entire gallery by aggregating their decisions. We first discuss the design to ensure that each compositor acquires distinct knowledge, then explain how compositors learn from each other to reduce their biases learned on noisy triplets. The batch-based classification loss is as follows:
\begin{equation}
\label{eq:lossbbc}
    \mathcal{L}_{BBC} = -\log\frac{\exp(\bm{\hat{g}_{}} \cdot \bm{g_{+})}}{\sum_{j=1}^{B}\exp(\bm{\hat{g}_{}} \cdot \bm{g_{j}} )},
\end{equation}
where $\bm{\hat{g}}$ is the composed feature from respective compositor, and $\bm{g_{j}}$ are candidates, among which the true match is $\bm{g_{+}}$.

\noindent\textbf{Pyramid Training for Image-Text Compositor.}\label{subsubsec:pyramid}
We develop a pyramid training paradigm for image-text compositors, which is inspired by the finding\!~\citep{lin2017feature,miech2021thinking} that the image features of high-resolution are semantically weak, while the image features of low-resolution are semantically strong. Through exploring the different spatial information of the reference image, the two image-text compositors $F^m_{IT}$ and $F^h_{IT}$ independently learn knowledge by leveraging the batch-based classification loss $\mathcal{L}^m_{IT}$ and $\mathcal{L}^h_{IT}$.
The independent batch-based classification loss makes each image-text compositor learn from the interactions between relative caption and different spatial information of the reference image, which enables these compositors to hold distinct knowledge from each other. 

\noindent\textbf{Auxiliary knowledge from Text-Image Compositor.}
The text-image compositor is a fancy component for generating the composed feature from the input, which is seldom referred to in previous works. It offers additional knowledge due to its distinct design from the image-text compositor. As discussed in Sec.\!~\ref{subsec:overview}, the text-image compositor views the data from another perspective, mainly focusing on the text-to-image retrieval with the reference image implying ``what to preserve'' conditioned on the text information, while the image-text compositor finds ``what to change'' in the reference image. We use two symmetric text-image compositors at the same depths of the image encoder, leveraging the batch-based classification loss $\mathcal{L}^m_{TI}$ and $\mathcal{L}^h_{TI}$.

\noindent\textbf{Collaborative Consensus Learning.} The triplet ambiguity problem leads to noisy triplets and biases the model learning. To mitigate this problem, we use the Kullback Leibler divergence loss (KL loss) for two image-text compositors. The KL loss promotes the compositors to learn from each other, reducing biases and reaching a consensus. This approach balances the preservation of distinct knowledge and the attainment of consensus. By enhancing cooperation and knowledge sharing, our method is more robust to the triplet ambiguity problem. 
Specifically, we denote the resulting posterior probability of $F_{IT}^m$ as $\bm{p^m}$ and that of $F_{IT}^h$ as $\bm{p^h}$. We set a target probability $\bm{p^w}$ as the weighted sum of both $\bm{p^m}$ and $\bm{p^h}$, which is given by:
\begin{equation}
    \bm{p^w} = \lambda_1 \cdot \bm{p^m} + \lambda_2 \cdot \bm{p^h},
\end{equation}
where $\lambda_{1}$ and $\lambda_{2}$ are weight coefficients, and the KL loss is formulated as:
\begin{equation}
\label{eq:losskl}
    \mathcal{L}_{KL} = D_{KL}(\bm{p^m}||\bm{p^w}) + D_{KL}(\bm{p^h}||\bm{p^w}),
\end{equation}
where $D_{KL}$ is the KL divergence distance. The KL loss reduces the biases of the compositors during training, which works alongside the batch-based classification loss in our approach. The preliminary experiments show that it is not essential to incorporate extra KL loss for the two text-image compositors. See Sec.\!~\ref{sec:discuss} for a detaileed explanation.
The final loss for training is the sum of the above loss functions:
\begin{equation}
    \mathcal{L} = \mathcal{L}^m_{IT}+\mathcal{L}^h_{IT}+\mathcal{L}^m_{TI}+\mathcal{L}^h_{TI}+\mathcal{L}_{KL},
\end{equation}
\textcolor{black}{where $L_{IT}^m$, $L_{IT}^h$, $L_{TI}^m$, and $L_{TI}^h$ are batch-based classification loss used for independently training each image-text/text-image compositor $F_{IT}^m$, $F_{IT}^h$, $F_{TI}^m$, and $F_{TI}^h$. The superscript $m$ indicates the mid-level input feature, while $h$ denotes the high-level feature. The subscript $IT$ indicates the image-text compositor, while the subscript $TI$ denotes the text-image compositor.}

\noindent\textbf{Joint Inference.}\label{subsubsec:joint_infer}
The four distinct compositors independently learn different knowledge from the data triplets and enable the knowledge transfer to reduce biases learned on noisy triplets. At the evaluation step, we involve each compositor in decision-making to further minimize individual bias. Specifically, we use each compositor to independently generate composed features and measure the similarity between any composed feature and target feature. The resulting similarity matrices are denoted as $P_{IT}^m$, $P_{IT}^h$, $P_{TI}^m$, $P_{TI}^h \in \mathbb{R}^{n_1 \times n_2}$, where $n_1$ and $n_2$ are the number of queries and target images in the gallery. The final similarity matrix for ranking the gallery is the weighted sum of four similarity matrices from distinct compositors:
\begin{equation} 
\label{eq:jointinfer}
    P = \alpha_1 \cdot P_{IT}^m + \alpha_2 \cdot P_{IT}^h + \alpha_3 \cdot P_{TI}^m + \alpha_4 \cdot P_{TI}^h,
\end{equation}
where $\alpha_{1} \dots \alpha_{4}$ are weight coefficients to balance the decisions from four compositors. Note that a common practice that concatenates multiple composed features as one query is a special case where all $\alpha$s are equal to $1$.



\section{Experiments} \label{experiments}

\subsection{Experimental Setup}
\label{subsec:implement}
\noindent\textbf{Datasets.} 
We evaluate Css-Net on three composed image retrieval datasets, \ie, Shoes\!~\citep{berg2010automatic}, FashionIQ\!~\citep{wu2021fashion}, and Fashion200k\!~\citep{vo2019composing}.

\begin{itemize}

    \item[$\bullet$] The Shoes dataset\!~\citep{berg2010automatic} is originally crawled from $like.com$ for attribute discovery. It is then annotated in the form of a triplet for dialog-based interactive retrieval. We follow VAL\!~\citep{chen2020image} to use $10,000$ training samples and $4,658$ evaluation samples.
    \item[$\bullet$] The FashionIQ dataset \!~\citep{wu2021fashion} is a language-based interactive fashion retrieval dataset with $77,684$ images across three categories: Dresses, Tops\&Tees, and Shirts. It includes $18,000$ triplets from $46,609$ training images, each containing a reference image, a target image, and two descriptive natural language captions. The evaluation procedure follows VAL \!~\citep{chen2020image} and CoSMo\!~\citep{lee2021cosmo}. 
    \item[$\bullet$] The Fashion200k dataset\!~\citep{han2017automatic} contains over $200k$ fashion images from various websites and is for attribute-based product retrieval. With descriptive attributes for each product, $172$k images are used for training and $33,480$ test queries for evaluation, following VAL and CoSMo methods. The relative descriptions are generated from attributes using an online-processing pattern. 
\end{itemize}




\begin{figure}[t]
    \centering
    \includegraphics[width=0.9\linewidth]{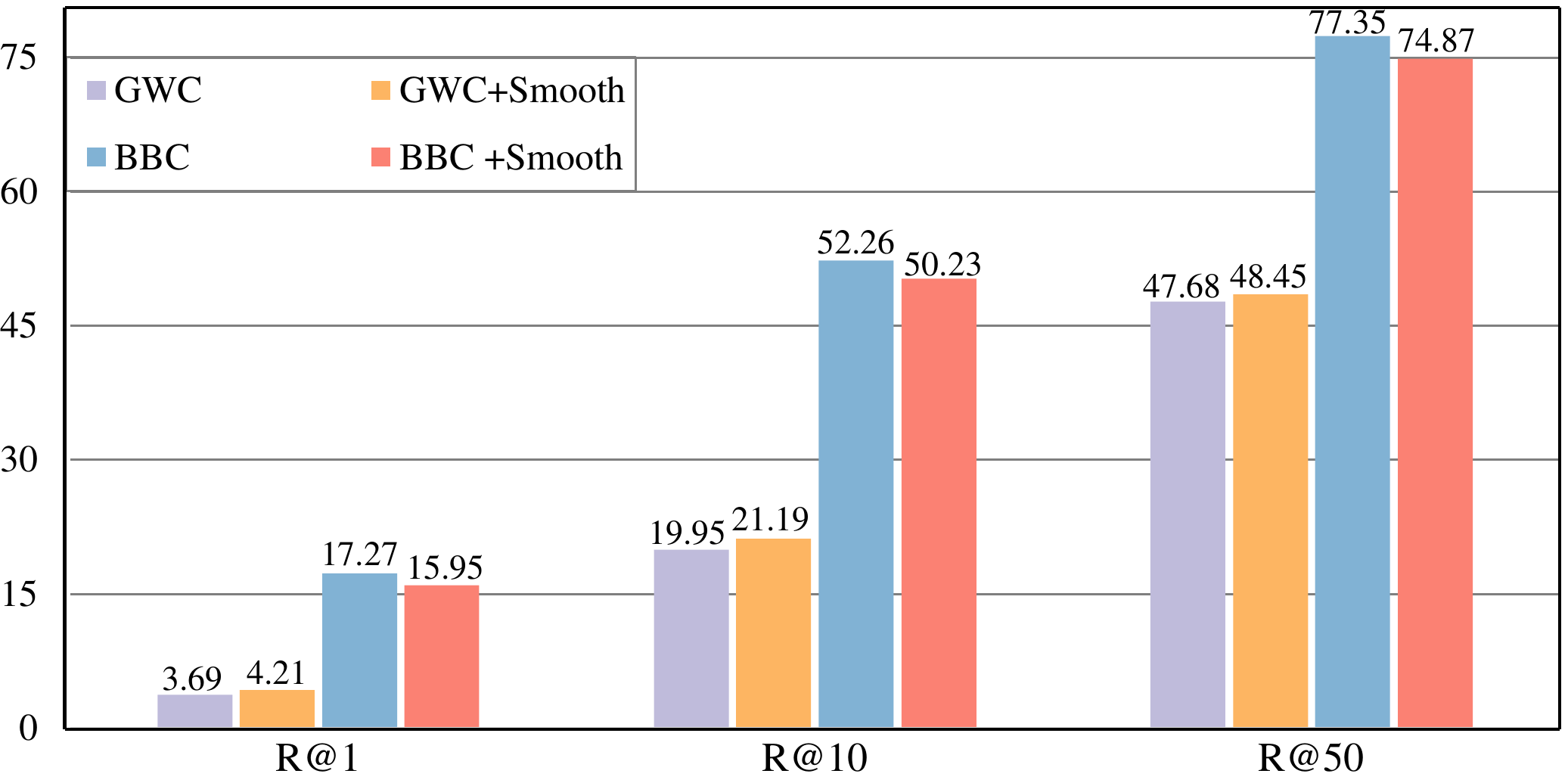}
    \caption{Comparison between the batch-based classification and the global-wise classification (GWC) on the Shoes dataset. GWC significantly degrades the performance since more false negative samples are involved due to triplet ambiguity.}
    \label{fig:twolosses}
\end{figure}

\subsection{Triplet Ambiguity Verification}
\label{subsec:ambiguity}

\noindent\textbf{Global-wise \emph{v.s.} Batch-based Optimization.}
To verify the negative impacts from the noisy triplets as shown in Fig.\!~\ref{fig:intro}, we quantitatively compare global-wise with batch-based optimization objectives. In particular, $\bullet$ Batch-based Classification (BBC): Limited negatives in the current batch are involved, and $\bullet$ Global-wise Classification (GWC): Mining more negative samples in the whole training set for comparison.

\begin{table*}[ht]
\centering
\setlength\tabcolsep{3pt}
  \resizebox{\linewidth}{!}{
\begin{tabular}{@{}lcccccccc}
\toprule
\multirow{2}{*}{Method} & \multicolumn{2}{c|}{Dress}        & \multicolumn{2}{c|}{Shirt}          & \multicolumn{2}{c|}{Toptee}         & \multicolumn{2}{c}{Average}        \\ \cmidrule(l){2-9} 
       & \multicolumn{1}{c}{R@10 $\uparrow$} & \multicolumn{1}{c|}{R@50 $\uparrow$} & \multicolumn{1}{c}{R@10 $\uparrow$} & \multicolumn{1}{c|}{R@50 $\uparrow$} & \multicolumn{1}{c}{R@10 $\uparrow$} & \multicolumn{1}{c|}{R@50 $\uparrow$} & \multicolumn{1}{c}{R@10 $\uparrow$} & \multicolumn{1}{c}{R@50 $\uparrow$} \\ \midrule
MRN\!~\citep{kim2016multimodal} & 12.32   & 32.18    & 15.88   & 34.33    & 18.11   & 36.33    & 15.44   & 34.28   \\
FiLM\!~\citep{perez2018film}   & 14.23   & 33.34    & 15.04   & 34.09    & 17.30   & 37.68    & 15.52  & 35.04   \\
TIRG\!~\citep{vo2019composing}   & 14.87   & 34.66    & 18.26   & 37.89    & 19.08   & 39.62    & 17.40   & 37.39   \\
VAL\!~\citep{chen2020image}    & 21.12   & 42.19    & 21.03   & 43.44    & 25.64   & 49.49    & 22.60    & 45.04   \\
DCNet\!~\citep{kim2021dual}  & 28.95   & 56.07    & 23.95   & 47.30     & 30.44   & 58.29    & 27.78   & 53.89   \\
CoSMo*\!~\citep{lee2021cosmo} & 26.45   & 52.43    & 26.94   & 52.99    & 31.95   & 62.09    & 28.45   & 55.84   \\
CLVC-Net$\dagger$ \!~\citep{wen2021comprehensive} & 29.85   & 56.47    & 28.75   & 54.76    & 33.50    & 64.00    & 30.70    & 58.41   \\
ARTEMIS\!~\citep{delmas2022artemis} &27.16   & 52.40    & 21.78   & 54.83    & 29.20   & 43.64    & 26.05   & 50.29   \\ 
MUR\!~\citep{chen2022composed} &30.60 & 57.46 & 31.54 & 58.29 & 37.37 & 68.41 & 33.17 & 61.39\\
CLIP4Cir\!~\citep{baldrati2022conditioned} &31.73 &56.02 &35.77 &57.02 &36.46 &62.77 &34.65 &58.60 \\ \midrule
Baseline &30.95 	&56.98	&31.48	&59.98  &36.97	&67.31		&33.13	&61.42 \\
Css-Net  & \textbf{33.65}   & \textbf{63.16}    & \textbf{35.96}   & \textbf{61.96}    & \textbf{42.65}   & \textbf{70.70}     & \textbf{37.42}   & \textbf{65.27}   \\ \bottomrule
\end{tabular}
}
\caption{Quantitative results on the FashionIQ dataset. The best results are in \textbf{bold}. The symbol * marks an updated results by the same authors. The symbol $\dagger$ indicates that this method deploys model ensemble (the same as below).}
\label{tab:fashioniq}
\end{table*}

If the data triplets do \textbf{NOT} have ambiguity, the global-wise classification has the potential to be comparable or even better since it uses more negative samples in the training set and potentially learns a better metric, which is consistent with many findings in metric learning\!~\citep{hermans2017defense,sheng2020mining,wang2020cross} and self-supervised learning\!~\citep{chen2020simple,he2020momentum}. Specifically, Given a query $q$ and features/prototypes $\{k_0, k_1, ...\}$ of candidate target images, where the true match is denoted as $k_+$. Two losses are given by:
\begin{equation}
    \mathcal{L}_{BBC} = -\log\frac{\exp(q\cdot k_+))}{\sum_{i=1}^{B}\exp(q\cdot k_i))}
\end{equation}
and
\begin{equation}
    \mathcal{L}_{GWC} =  -\log\frac{\exp(q\cdot k_+))}{\sum_{i=1}^{N}\exp(q\cdot k_i))},
\end{equation}
where $B$ is the batch size, and $N$ is the number of IDs (classes) in the training set. The only difference between them is that $\mathcal{L}_{GWC}$ involves more negative counterparts, which results in high false negative rates if the triplet ambiguity does exist. We conduct experiments on the Shoes dataset\!~\citep{berg2010automatic} using two losses, respectively, under the same settings of CoSMo\!~\citep{lee2021cosmo}. We observe that batch-based methods outperform global-wise methods by a large margin, as shown in Fig.\!~\ref{fig:twolosses}. The experimental results confirm our triplet ambiguity assumption: the training data contains many noisy triplets (\ie, false negative samples). Although batch-based classification suffers less from triplet ambiguity, the single compositor still faces some noisy negative triplets in the batch and produces a sub-optimal solution.

\noindent\textbf{Label Smoothing.}
\label{subsubsec:smoothing}
One intuitive way we consider to alleviate the triplet ambiguity problem is label smoothing. The motivation is that there are many false negative samples due to the triplet ambiguity problem, and label smoothing could alleviate the overfitting to the annotated true match. In label smoothing, the label $\bm{y} = [y_1, \dots y_n]$ is not a hard one-hot label rather than a soft one-hot label, which is given by: 
\begin{equation}
y_i = 
\begin{cases}
&1~(if~i =  c) \\
&0~(if~i\ne c)
\end{cases}
\Longrightarrow 
y_i = 
\begin{cases}
&1-\varepsilon ~(if~i =  c) \\
&\frac{\varepsilon}{B-1} ~(if~i\ne c),
\end{cases}
\end{equation}
where $y_i$ is the label for class $i$, $c$ is the corresponding class of the query, $B$ is the batch size, and $\varepsilon$ is a hyperparameter for label smoothing and is set to be $0.1$. We use label smoothing for both the batch-based classification and the global-wise classification, which are presented in Fig.\!~\ref{fig:twolosses}. The experimental results indicate that label smoothing deteriorates the performance of batch-based classification but enhances the performance of global-wise classification. This is because $\bullet$~global-wise classification is severely affected by triplet ambiguity due to high false negative rate, while batch-based classification is affected only when noisy negative triplets are in the batch; $\bullet$~Label smoothing could relieve the triplet ambiguity but introduce another problem that many true negative target samples are assigned weights, which impairs the model training for batch-based classification. The experimental results also verify the effectiveness of KL loss as another form of soft label.

\begin{table}[t]
\small
\setlength\tabcolsep{25pt}
\resizebox{\linewidth}{!}{
\begin{tabular}{@{}lccc}
\toprule
\multirow{2}{*}{Method} & \multicolumn{3}{c}{Shoes} \\ \cmidrule(l){2-4} 
       & R@1 $\uparrow$    & R@10 $\uparrow$  & R@50 $\uparrow$  \\ \midrule
MRN\!~\citep{kim2016multimodal}    & 11.74    & 41.70  & 67.01  \\
FiLM\!~\citep{perez2018film}   & 10.19    & 38.89  & 68.30  \\
TIRG\!~\citep{vo2019composing}   & 12.60    & 45.45  & 69.39  \\
VAL\!~\citep{chen2020image}    & 16.49   & 49.12  & 73.53  \\
CoSMo\!~\citep{lee2021cosmo}  & 16.72   & 48.36  & 75.64  \\
DCNet\!~\citep{kim2021dual}  & -       & 53.82  & 79.33  \\
CLVC-Net$\dagger$\!~\citep{wen2021comprehensive}  & 17.64   & 54.39  & 79.47  \\
MUR\!~\citep{chen2022composed} & 18.41 & 53.63 & 79.84 \\
ARTEMIS\!~\citep{delmas2022artemis} & 18.72   & 53.11  & 79.31  \\ \midrule
Baseline                & 17.27   &52.26   &77.35   \\
Css-Net& \textbf{20.13}   & \textbf{56.81}  & \textbf{81.32}  \\ \bottomrule
\end{tabular}
}
\caption{Quantitative results on the Shoes dataset. \textcolor{black}{The best results are in \textbf{bold}. The symbol $\dagger$ indicates that this method deploys model ensemble. The proposed method has achieved competitive performances in all three metrics R@1, 10, 50.}}
\label{tab:shoes}
\end{table}
\begin{table}[t]
\setlength\tabcolsep{27pt}
\resizebox{\linewidth}{!}{
\begin{tabular}{@{}lccc}
\toprule
\multirow{2}{*}{Method} & \multicolumn{3}{c}{Fashion200k} \\ \cmidrule(l){2-4} 
       & R@1 $\uparrow$      & R@10 $\uparrow$    & R@50 $\uparrow$    \\ \midrule
MRN\!~\citep{kim2016multimodal}    & 13.4    & 40.0  & 61.9  \\
FiLM\!~\citep{perez2018film}   & 12.9    & 39.5  & 61.9  \\
TIRG\!~\citep{vo2019composing}   & 14.1      & 42.5     & 63.8     \\
VAL\!~\citep{chen2020image}    & 21.2      & 49       & 68.8     \\
DCNet\!~\citep{kim2021dual}  & -         & 46.9     & 67.6     \\
CoSMo\!~\citep{lee2021cosmo}  & 23.3      & 50.4     & 69.3     \\
CLVC-Net$\dagger$\!~\citep{wen2021comprehensive}                & 22.6      & \textbf{53.0}      & \textbf{72.2}     \\
ARTEMIS\!~\citep{delmas2022artemis}     & 21.5      & 51.1     & 70.5     \\ \midrule
Baseline                & 20.9      & 47.7     & 67.8     \\
Css-Net& 22.2      & 50.5     & 69.7                        \\
Css-Net$\dagger$ & \textbf{23.4}     & 52.0     & 72.0                       \\ \bottomrule
\end{tabular}
} 
\caption{Quantitative results on the Fashion200k dataset. \textcolor{black}{The best results are in \textbf{bold}. The symbol $\dagger$ indicates that this method deploys model ensemble. The proposed method has achieved competitive performances.}}
\label{tab:fashion200k}
\end{table}

\begin{table}[t]

\small

\end{table}

\subsection{The Effectiveness of Our Method}
\label{subsec:effective}

We present the experimental results in Table\!~\ref{tab:fashioniq}, Table\!~\ref{tab:shoes}, and Table\!~\ref{tab:fashion200k}. 
We could make two observations: \textbf{(1) We adopt a competitive baseline with few modifications.} As mentioned in Sec.\!~\ref{subsec:implement}, we adopt the CoSMo as our baseline and replace the LSTM with a more robust text encoder: RoBERTa, and observe consistent improvement. For example, on the FashionIQ dataset, our baseline improves CoSMo by $4.68\%$ R@10 on average and surpasses CoSMo by $3.90\%$ R@10 on the Shoes dataset. We infer that RoBERTa is more robust than LSTM\!~\citep{hochreiter1997long} to accurately capture the textual information. However, our baseline is slightly lower than the reported results of CoSMo on Fashion200k, as the authors do not provide sufficient implementation details for reproducing. This also limits comparing our method with CQBIR\!~\citep{zhang2022comprehensive}, whose baseline uses faster RCNN\!~\citep{girshick2015fast} as a different image encoder. Nevertheless, our method is more effective than CQBIR on FashionIQ and Shoes, where the triplet ambiguity problem is more serious.
\textbf{(2) The proposed Css-Net could further improve and advances the state of the art on such a strong baseline, verifying the effectiveness of Css-Net.} For example, Table\!~\ref{tab:fashioniq} shows Css-Net improves retrieval accuracy on all FashionIQ subsets. Compared to the baseline, it gains $+2.70\%$ R@10 on Dress, $+4.48\%$ R@10 on Shirt, and $+5.68\%$ R@10 on TopTee. Compared to previous works, our method brings overall improvements (e.g., $+2.77\%$ R@10 and $+6.67\%$ R@50 on average by CLIP4Cir). The improvements are significant and empirically validate the effectiveness of Css-Net for handling the triplet ambiguity problem. Besides in Table\!~\ref{tab:shoes}, Css-Net surpasses the state-of-the-art (CLVC-Net) on the Shoes dataset, achieving improvements of $+2.49\%$ R@1 and $+2.42\%$ R@10, which further demonstrates that Css-Net is robust across different datasets. Table\!~\ref{tab:fashion200k} presents Fashion200k results. Although our baseline is below the reported results of CosMo because of insufficient implementation details for reproduction, Css-Net brings a considerable improvement (\eg, $+2.8\%$ R@10 over the baseline ) and is still competitive with many SOTA works especially when applying the model ensemble (\eg, $+4.3\%$ R@10).

\subsection{Diagnostic Experiments}
\label{subsec:diagnostic}

\begin{table}[t]
\setlength\tabcolsep{37pt}

  \resizebox{\linewidth}{!}{
\begin{tabular}{@{}lccc}
\toprule
\multirow{2}{*}{Method} & \multicolumn{3}{c}{Shoes} \\ \cmidrule(l){2-4} 
       & R@1 $\uparrow$     & R@10 $\uparrow$  & R@50 $\uparrow$  \\ \midrule
$F_{IT}^h$  & 17.27   &52.26   &77.35   \\
$F_{IT}^l+F_{IT}^h$    & 18.24 & 52.14  & 78.12  \\ 
$F_{IT}^l+F_{IT}^m+F_{IT}^h$   & 18.81   & 54.21  & 79.55  \\ 
$F_{IT}^m+F_{IT}^h$    & 19.10   &54.69   &79.63   \\
\bottomrule
\end{tabular}
}
\caption{Comparison of various pyramid training methods on the Shoes dataset. These methods are trained and evaluated independently. \textcolor{black}{$F_{IT}^l$, $F_{IT}^m$, and $F_{IT}^h$ represent the low-level, mid-level, and high-level image-text compositor, respectively. The low-level compositor is useful, whereas the mid and high-level features show better performance.}}
\label{tab:pyramid}

\end{table}

\begin{table}[t]
\centering
  \setlength\tabcolsep{15pt}
  
  \resizebox{\linewidth}{!}{
\begin{tabular}{@{}lccccc}
\toprule
\multirow{2}{*}{$\mathcal{L}_{IT}^m$} &  \multirow{2}{*}{$\mathcal{L}_{TI}^h+L_{TI}^m$} &  \multirow{2}{*}{$\mathcal{L}_{KL}$}  & \multicolumn{3}{c}{Shoes} \\ \cmidrule(l){4-6} 
& & & R@1 $\uparrow$     & R@10 $\uparrow$  & R@50 $\uparrow$  \\ \midrule
\multicolumn{2}{l}{\hspace*{-3.5pt}Baseline: (only $\mathcal{L}_{IT}^h$}) &  & 17.27    & 52.26  & 77.35  \\ \midrule
$\bm{\checkmark}$& &     & 19.10\textcolor{green!70!black}{(+1.83)}  & 54.69\textcolor{green!70!black}{(+2.43)}  & 79.63\textcolor{green!70!black}{(+2.28)}  \\ 
$\bm{\checkmark}$&$\bm{\checkmark}$ &   & 19.47\textcolor{green!70!black}{(+2.20)}   & 54.63\textcolor{green!70!black}{(+2.37)}  & 80.46\textcolor{green!70!black}{(+3.11)}  \\
$\bm{\checkmark}$&$\bm{\checkmark}$ & $\bm{\checkmark}$  & 20.13\textcolor{green!70!black}{(+2.86)}       & 56.81\textcolor{green!70!black}{(+4.55)}  & 81.32\textcolor{green!70!black}{(+3.97)}  \\ \bottomrule
\end{tabular}
}
\caption{Efficacy of model designs. \textcolor{black}{$L_{IT}^m$, $L_{IT}^h$, $L_{TI}^m$, and $L_{TI}^h$ are batch-based classification loss defined in Eqn.\!~\ref{eq:lossbbc}, and $L_{KL}$ is the KL loss defined in Eqn.\!~\ref{eq:losskl}.}}
  \label{tab:diagnostic}
\end{table}

\begin{table}[t]
\setlength\tabcolsep{35pt}

  \resizebox{\linewidth}{!}{
\begin{tabular}{@{}lccc}
\toprule
\multirow{2}{*}{Inference Method} & \multicolumn{3}{c}{Shoes} \\ \cmidrule(l){2-4} 
       & R@1 $\uparrow$     & R@10 $\uparrow$   & R@50 $\uparrow$   \\ \midrule
$F_{IT}^m$    & 15.72   &51.17   &78.89   \\
$F_{IT}^h $   & 18.35  & 55.15            &80.52   \\
$F_{TI}^m$     & 17.06  & 53.35            &78.92    \\
$F_{TI}^h$     & 16.58  & 52.17            &77.77    \\ \midrule
Joint Inference (Eq.\!~\ref{eq:jointinfer})     & \textbf{20.13}  & \textbf{56.81}            &\textbf{81.32}    \\
\bottomrule
\end{tabular} }
\caption{Effect of joint inference. We train Css-Net with four compositors on Shoes once and separately evaluate each compositor. \textcolor{black}{Joint inference refers to using the weighting scheme (Eqn.\!~\ref{eq:jointinfer}) to combine decisions from all the compositors .}}
\label{tab:joint_infer}
\end{table}

\begin{table}[t]
\centering
\setlength\tabcolsep{18pt}

\resizebox{\linewidth}{!}{%
\begin{tabular}{@{}lccc@{}}
\toprule
              & Total time (s) $\downarrow$ & Time per query (ms) $\downarrow$ & Time per target (ms) $\downarrow$ \\ \midrule
Baseline (one) & 168.2          & 50.2                & 56.4            \\
Css-Net (four)   & 195.8          & 58.5                & 65.7            \\ \bottomrule
\end{tabular}%
}
 \caption{Inference time cost for the baseline and Css-Net. \textcolor{black}{Total time refers to the time taken to process all queries. Time per query indicates the average time spent on each query, while time per target represents the average time used to process each target in the gallery. }}
\label{tab:latency}
\end{table}

\noindent\textbf{Pyramid Training.}
\label{subsec:layer2}
In Sec.\!~\ref{subsubsec:pyramid}, we present the design of the pyramid training, which exploits the image features from the mid-level and high-level blocks of the image encoder. We verify its effectiveness by comparing it with different designs. Table\!~\ref{tab:pyramid} reports the experimental results. Our baseline is $F_{IT}^m+F_{IT}^h$ used in Css-Net. We conduct experiments on two variants for pyramid training: 1) $F_{IT}^l+F_{IT}^h$, which uses the image features from block2 and block4 of the ResNet, and 2) $F_{IT}^l+F_{IT}^m+F_{IT}^h$, utilizing three image-text compositors at three depths. Both variants perform worse than Css-Net, e.g., $-2.55\%$ and $-0.48\%$ on the R@10 metric. However, they both surpass $F_{IT}^h$ using only one image-text compositor at block4. These results indicate that 1) the low-level image feature is too semantically weak to provide image information, and 2) groups perform better than individuals. 

\noindent\textbf{Efficacy of Model Designs.}
Table\!~\ref{tab:diagnostic} shows the effectiveness of our core idea, which uses four different compositors with KL loss to relieve the triplet ambiguity problem. We make three observations from the table. First, employing image-text compositors at other layers of the image encoder (\ie, $\mathcal{L}_{IT}^m$) can mitigate the triplet ambiguity problem and improve the performance significantly ($77.35\% \rightarrow 79.63\%$ at R@50 metric). This indicates that two image-text compositors can benefit from the interactions between the relative caption and different spatial information of the reference image. Second, adding a new compositor module, text-image compositor, to this task (\ie, $\mathcal{L}_{TI}^m+\mathcal{L}_{TI}^h$) can further improve the performance ($79.63\% \rightarrow 80.46\%$ at R@50 metric). This demonstrates the advantage of auxiliary knowledge. Third, applying an extra KL loss for two image-text compositors ($\mathcal{L}_{KL}$) can enhance the performance notably ($80.46\% \rightarrow 81.32\%$ at R@50 metric). This suggests that the KL loss enables two image-text compositors to share their knowledge, thus minimizing the biases learned from noisy triplets.

\noindent\textbf{Effect of Joint Inference}
At the evaluation stage, Css-Net allows compositors to jointly make the decision as introduced in Sec.\!~\ref{subsubsec:joint_infer}. As shown in Table\!~\ref{tab:joint_infer}, joint inference surpasses single compositor and verifies our motivation that groups perform better than individuals and could be used to reduce their own prediction biases mainly caused by the triplet ambiguity problem. 

\noindent\textbf{Computational cost at inference}
Css-Net uses four compositors that share the same image and text encoders, thus adding minimal retrieval latency. The inference time is shown in Table\!~\ref{tab:latency}, ranging from loading the model to displaying results. The experiments are conducted with GeForce RTX 2080 Ti, using $33,480$ queries and $29,789$ targets.

\noindent\textbf{Implementation Details.} 
We modify CoSMo\!~\citep{lee2021cosmo} as our baseline by replacing LSTM\!~\citep{graves2012long} with RoBERTa\!~\citep{liu2019RoBERTa} as the text encoder. ResNet-50\!~\citep{he2016deep} serves as the image encoder for Shoes and FashionIQ datasets, while ResNet-18\!~\citep{he2016deep} is used for Fashion200k. Embedding space dimension $C$ is $512$. Text feature shape is $C_{in}' \times L$, with $C_{in}'$ being $768$ and $L$ is the sentence length. During training, we set $\lambda_1 = 10$ and $\lambda_2 = 1$, while evaluation uses $\alpha_{1} \dots \alpha_4 = 1, 0.5, 0.5, 0.5$. We adopt the standard evaluation metric in retrieval, \ie, Recall@K, denoted as R@K for short. We use a random seed for each experiment and repeat it five times for the final results.
we employ the Adam optimizer\!~\citep{kingma2014adam} with $\beta_1 = 0.9$ and $\beta_2 = 0.999$. On Shoes and FashionIQ, the batch size is set to be $32$ and the base learning rates of the text encoder and other modules are $2e-6$ and $2e-5$, respectively. On Fashion200k, the batch size is set to be $128$ and the base learning rates are $2e-6$ and $2e-4$, respectively. We adopt warm-up for the first $5$ epochs, decay learning rate by $10$ at epochs $35$ and $45$ during training. The total training epoch is $50$.

\section{Further Analysis and Discussion}
\subsection{Further Qualitative Analysis}
\label{sec:qualitative}
\begin{figure}[t]
    \centering
    \includegraphics[width=\linewidth]{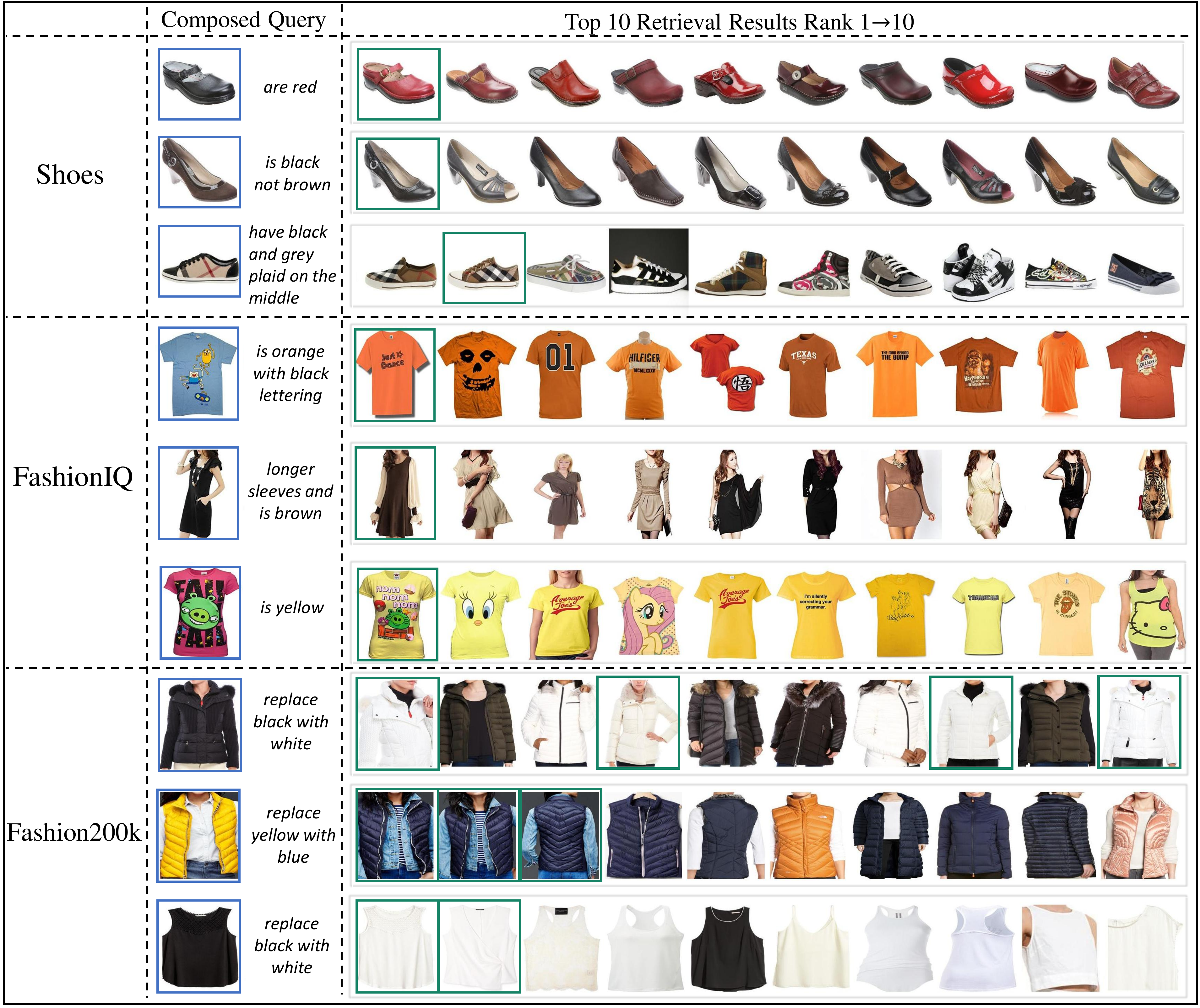}
    \caption{Top-10 retrieval results on three datasets. The composed queries consist of a reference image and a relative caption that describes the desired modification. The \textcolor{blue}{blue}/\textcolor{green!70!black}{green} boxes refer to the reference image and the true match(es).}
    \label{fig:top10}
\end{figure}
Fig.\!~\ref{fig:top10} shows the top-10 retrieval results on three datasets: Shoes, Fashion200K, and FashionIQ. We make three key observations from these results: (1) Css-Net can capture the information of the reference image and the relative caption for both coarse-grained and fine-grained queries. For example, the first query of Shoes and the third query of FashionIQ retrieve the correct matches easily, and the first and second queries of FashionIQ also find the correct matches. These queries have clear and distinctive features that can be matched by Css-Net. (2) The model sometimes fails to retrieve the correct matches due to the triplet ambiguity problem, \eg, the first query of Fashion200K retrieves some negative samples but are still highly related to the query. (3) Css-Net is less sensitive to some detailed information such as location. For example, the third query in Shoes retrieves a shoe that is visually similar but has a wrong paid location, because the dataset has few similar training samples. Improving the sensitivity of the model to the detailed information is a direction for our future work. We plan to explore more fine-grained features to enhance Css-Net in the future works.

\subsection{Comparison with Most Relevant Works}
\label{sec:relevant}

We compare our Css-Net with VAL\!~\citep{chen2020image} and CLVC-Net\!~\citep{wen2021comprehensive}, which are most relevant to our work.

(1) Our Css-Net differs from the hierarchical matching strategies in VAL:
\begin{itemize}
    \item Our Css-Net facilitates knowledge sharing between compositors at various depths for consensus, instead of independent learning in VAL. 
    \item Our Css-Net observes that the low-level compositor does not contribute to collaborative learning and omits it to enhance the recall performance and efficiency (\textit{See Table 1}). 
    \item Our Css-Net implements an adjustable weighted sum during evaluation, enabling individuals to make decisions as a group.
\end{itemize}

(2) Our Css-Net differs from the model ensemble design in CLVC-Net: 
\begin{itemize}
    \item Our Css-Net is more efficient since all compositors share the same encoder stem (Table 7), while model ensembling in CLVCNet employs several independent backbones.
    \item Our Css-Net encourages intra-modal and inter-modal knowledge sharing via collaborative learning between compositors, while model ensembling does not entail additional loss or learning among the models.
    \item Our Css-Net acknowledges that the compositors have different knowledge and thus assign adaptive weights, while model ensembling usually presumes that the models are independent and equally important.
    \item Our Css-Net enables single compositor to perform better could further benefits from model ensembling (See Table 6), while model ensembling  does not improve single compositor prediction.
\end{itemize}

\subsection{Discussion of Collaborative Learning}
\label{sec:discuss}
\begin{figure}[h!]
    \centering
    \includegraphics[width=.65\linewidth]{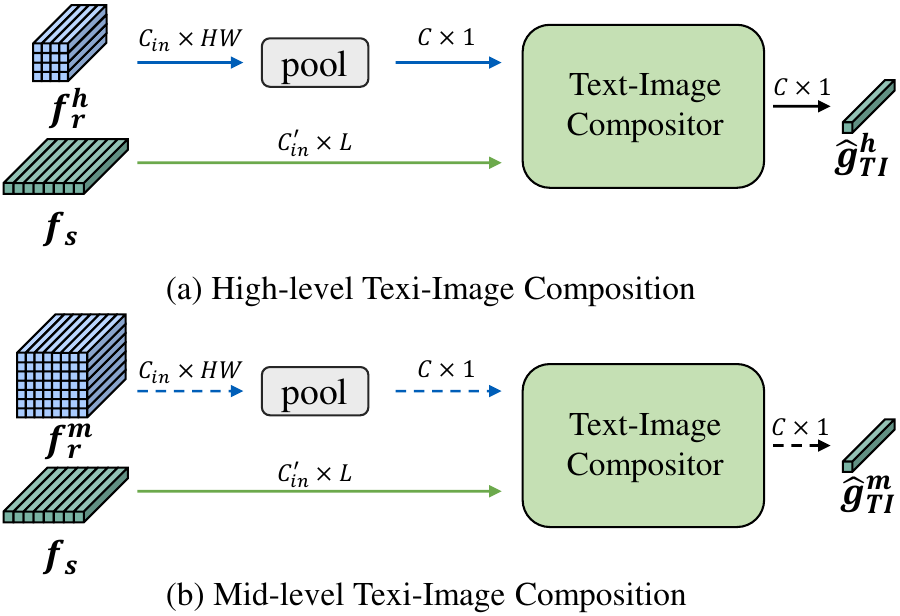}
    \caption{A brief illustration of two text-image compositors with the input shape. Please refer to Fig.\!~\ref{fig:overview} for the entire framework.}
    \label{fig:suppl-kl}
\end{figure}

We apply a KL loss between text-image compositors in a preliminary experiment, but find that it is not as significant as the KL loss between image-text compositors. This is because the inputs for the text-image compositors are too similar, as shown in Fig.\!~\ref{fig:suppl-kl}. Specifically, both text-image compositors receive a pooled reference image feature with identical dimensions and share the same text representations. Therefore, the main function of these text-image compositors is to act as auxiliary decision-makers during joint inference, addressing the triplet ambiguity issue. For simplicity and efficiency, we do not incorporate additional KL loss for the text-image compositors. However, we note that the text-image compositors still play an important role in our framework, as they provide complementary information to the image-text compositors and improve the retrieval performance.

\subsection{Analysis for hyperparameters}
\label{sec:hyperparameter}

\begin{table}[h!]
\centering
\setlength\tabcolsep{40pt}

\resizebox{\linewidth}{!}{%
\begin{tabular}{@{}lccc}
\toprule
                                   & R@1   & R@10  & R@50  \\ \midrule
Css-Net ($\alpha_{1-4} = 1,1,1,1$)            & 20.04 & 56.44 & 80.87 \\ 
Css-Net ($\alpha_{1-4} = 1,0.5,0.5,0.5$) & 20.13 & 56.81 & 81.32 \\ \bottomrule
\end{tabular}%
}
\caption{Ablation for hyperparameter $\alpha$.}
\label{tab:alpha}
\end{table}

\begin{table}[h!]
\centering
\setlength\tabcolsep{40pt}

\resizebox{\linewidth}{!}{%
\begin{tabular}{@{}lccc}
\toprule
                             & R@1   & R@10  & R@50  \\ \midrule
Css-Net ($\lambda_{1-2} = 1,1$)     & 19.95 & 56.64 & 80.55 \\ 
Css-Net ($\lambda_{1-2} = 10,1$) & 20.13 & 56.81 & 81.32 \\ \bottomrule
\end{tabular}%
}
\caption{Ablation for Hyperparameter $\lambda$.}
\label{tab:lambda}
\end{table}

In this work, the hyperparameters $\alpha$s and $\lambda$s are not handpicked, as we empirically find that they are not sensitive and do not affect the model performance significantly. We set $\alpha$s to be 1\!~:\!~0.5\!~:\!~0.5\!~:\!~0.5 based on the observation that the high-level image-text compositor performs best among all compositors (Table~\ref{tab:alpha}) and we want this compositor to act like a leader in the group. Similarly, we use $\lambda$s $=1$ for all compositors, as we have some preliminary experiments that show similar results with this setting (Table\!~\ref{tab:lambda}). To demonstrate this, we add some experimental results on the Shoes dataset, which is another challenging benchmark for composed image retrieval. The results show that our Css-Net achieves competitive performance with different values of $\alpha$s and $\lambda$s, indicating that our model is robust and stable to the choice of hyperparameters.

\textcolor{black}{\subsection{Effect of More Annotation Noise}}

\textcolor{black}{In this work, we aim to relieve the issue of noisy annotations, which can compromise the entire training process. Further, we artificially increased the noise intensity during training by manually manipulating relevant captions, such as random deletion, random swap, and random insertion proposed in a NLP work \citep{wei2019eda}. To be more specific, we conducted an experiment on the Shoes dataset for both the baseline and Css-Net. For each relative caption, there is a 50\% probability of adding one of three types of noise: Each word in the sentence has a 50\% probability of being deleted; half of the words in the sentence are replaced with synonyms; and new words are inserted into half of the word intervals. The performance of the newly developed baseline and Css-Net are shown in Table\!~\ref{tab:noise}.}

\begin{table}[h!]
\centering
\setlength\tabcolsep{40pt}

\resizebox{\linewidth}{!}{%
\begin{tabular}{@{}lccc}
\toprule
Method                             & R@1   & R@10  & R@50  \\ \midrule
Baseline (\textit{w} noise)     & 16.29 & 50.14 & 75.91 \\ 
Css-Net (\textit{w} noise) & 19.07 & 55.69 & 78.98 \\ \midrule
Baseline (\textit{w/o} noise)     & 17.27 & 52.26 & 77.35 \\ 
Css-Net (\textit{w/o} noise) & 20.13 & 56.81 & 81.32 \\ 
\bottomrule
\end{tabular}%
}
\caption{Effect of annotation noise (w/o refers to without; w refers to with).}
\label{tab:noise}
\end{table}

\section{Conclusion} \label{conclusion}
We present a Consensus Network (Css-Net) for composed image retrieval. Css-Net aims to relieve the inherent triplet ambiguity problem, which arises when the dataset contains multiple false-negative candidates that match the same query. This problem stems from annotators describing only simple properties and frequently overlooking fine-grained details of the images. The resulting noisy triplets significantly compromise the metric learning objective and bias the single compositor. To this end, Css-Net employs a consensus module with four compositors that possess distinct knowledge. As a group, compositors learn mutually when training and infer collaboratively during evaluation, effectively minimizing the negative effects caused by the triplet ambiguity problem.
Extensive experiments show that Css-Net has achieved competitive recall performance on three widely-used benchmarks, without substantially increasing the inference time.


\bibliographystyle{elsarticle-num-names} 
\bibliography{refs}

\end{document}